\documentclass{article}

\usepackage[preprint]{neurips_2025}

\usepackage[utf8]{inputenc} % allow utf-8 input
\usepackage[T1]{fontenc}    % use 8-bit T1 fonts
\usepackage[colorlinks,     % colorful hyperlinks
            linkcolor=citecolor,
            anchorcolor=blue,
            citecolor=citecolor,   
            ]{hyperref}
\usepackage{url}            % simple URL typesetting
\usepackage{booktabs}       % professional-quality tables
\usepackage{amsfonts}       % blackboard math symbols
\usepackage{nicefrac}       % compact symbols for 1/2, etc.
\usepackage{microtype}      % microtypography
\usepackage{xcolor}         % colors

\usepackage{algorithm}
\usepackage{algorithmic}
\usepackage[utf8]{inputenc} % allow utf-8 input
\usepackage[T1]{fontenc}    % use 8-bit T1 fonts
\usepackage{url}            % simple URL typesetting
\usepackage{booktabs}       % professional-quality tables
\usepackage{amsfonts}       % blackboard math symbols
\usepackage{nicefrac}       % compact symbols for 1/2, etc.
\usepackage{microtype}      % microtypography
\usepackage{xcolor}         % colors
\usepackage{amsmath}
\usepackage{amssymb}
\usepackage{mathtools}
\usepackage{amsthm}
\usepackage{multirow}
\usepackage{tabularx}
\usepackage{microtype}
\usepackage{graphicx}
\usepackage{subfigure}
\usepackage{pifont}
\usepackage{empheq}
\usepackage{amssymb}
\usepackage{pifont}%
\usepackage{colortbl}
\usepackage{cases}  
\usepackage{bm}
\usepackage[hypcap=false]{caption}  % 用于\captionof命令
\usepackage{array}    % 用于表格格式控制
\usepackage{booktabs} % 用于专业表格线
\usepackage{xcolor}   % 用于单元格颜色
\usepackage{adjustbox} % 用于对齐调整

\definecolor{brickred}{rgb}{0.8, 0.25, 0.33}
\definecolor{bostonuniversityred}{rgb}{0.8, 0.0, 0.0}
\definecolor{brightmaroon}{rgb}{0.76, 0.13, 0.28}
\definecolor{candyapplered}{rgb}{1.0, 0.03, 0.0}
\definecolor{carminered}{rgb}{1.0, 0.0, 0.22}
\definecolor{coralred}{rgb}{1.0, 0.25, 0.25}
\definecolor{cornellred}{rgb}{0.7, 0.11, 0.11}
\definecolor{citecolor}{RGB}{17,80,197}
\definecolor{myblue}{RGB}{0, 102, 204}

\newtheorem{theorem}{Theorem}

\newcommand{\diag}{\mathrm{diag}}
\newcommand{\tr}{\mathrm{tr}}

\newcommand{\suchthat}{\mathrm{s.t.}}

\DeclareMathOperator*{\argmin}{argmin}

\newlength\savewidth

\title{SPAP: Structured Pruning via Alternating Optimization and Penalty Methods}

\author{%
  Hanyu Hu \\
  Department of Mathematics \\
  The University of Hong Kong \\
  \texttt{hhy1224@connect.hku.hk} \\
  \And
  Xiaoming Yuan\thanks{Corresponding author} \\
  Department of Mathematics \\
  The University of Hong Kong \\
  \texttt{xmyuan@hku.hk} \\
}

\begin{document}

\maketitle

\begin{abstract}

The deployment of large language models (LLMs) is often constrained by their substantial computational and memory demands. While structured pruning presents a viable approach by eliminating entire network components, existing methods suffer from performance degradation, reliance on heuristic metrics, or expensive finetuning. To address these challenges, we propose SPAP (Structured Pruning via Alternating Optimization and Penalty Methods), a novel and efficient structured pruning framework for LLMs grounded in optimization theory. SPAP formulates the pruning problem through a mixed-integer optimization model, employs a penalty method that effectively makes pruning decisions to minimize pruning errors, and introduces an alternating minimization algorithm tailored to the splittable problem structure for efficient weight updates and performance recovery. Extensive experiments on OPT, LLaMA-3/3.1/3.2, and Qwen2.5 models demonstrate SPAP's superiority over state-of-the-art methods, delivering linear inference speedups (1.29$\times$ at 30\% sparsity) and proportional memory reductions. Our work offers a practical, optimization-driven solution for pruning LLMs while preserving model performance.
        
\end{abstract}

\section{Introduction}
The rapid advancement of large language models (LLMs) has revolutionized natural language processing~\citep{zhang2022opt,openai2023gpt,touvron2023llama,touvron2023llama2, meta2023llama3,team2023gemini,grattafiori2024llama3-1,apple2023, deepseekai2025deepseekv3technicalreport,qwen2025qwen25technicalreport}, yet their enormous size poses significant challenges for practical deployment. Pruning techniques have emerged as a cornerstone for LLM compression~\citep{frantar2023sparsegpt, sun2023simple, ma2023llm_pruner, shen2024search, fang2024maskllm}, offering a promising solution to reduce model size and computational requirements while preserving performance. While unstructured pruning~\citep{frantar2023sparsegpt,sun2023simple, dong2024pruner, zhao2024convex} provides flexibility by zeroing out individual weights, it relies on specialized hardware for sparse computations and often yields marginal speedups due to sparsity patterns. Similarly, semi-structured pruning methods~\citep{holmes2021nxmtransformer,meng2024alps,fang2024maskllm} depend on specialized kernels and hardware supports like NVIDIA's 2:4 pattern~\citep{mishra2021accelerating}, making them highly hardware-dependent and thus limiting their applicability. Empirical studies further show that semi-structured pruning underperforms structured methods in inference acceleration at equivalent sparsity levels~\citep{ashkboos2024slicegpt}. In contrast, structured pruning distinguishes itself by eliminating entire network components (e.g., channels, heads, or layers) rather than individual weight entries, enabling direct computational speedups and broad hardware compatibility without relying on specialized hardware or sparse computations. These advantages make structured pruning a robust and hardware-agnostic strategy for LLM compression.

Despite its compelling advantages, existing structured pruning methods face three fundamental challenges when applied to modern LLM architectures. 
First, removing larger structural components inherently leads to greater performance degradation, making preservation of model quality particularly difficult~\citep{men2024shortgpt,yang2024laco, zhu2024survey}. 
Second, modern architectures like LLaMA-3~\citep{meta2023llama3} and Qwen2.5\citep{qwen2025qwen25technicalreport} employ grouped query attention (GQA)~\citep{ainslie2023gqa}, which reduces attention layer parameters to just 19.23\% and 12.60\% of decoder layers respectively through weight sharing, while multi-layer perceptron (MLP) layers dominate with 80.77\% and 87.40\% parameter shares. 
This architectural shift means approaches that focus on attention heads yield diminishing returns and may cause significant performance drops~\citep{hu2025fasp,wang2024cfsp}. 
Third, current methods exhibit notable limitations: LLM-Pruner~\citep{ma2023llm_pruner} requires extensive finetuning for recovery, while SliceGPT~\citep{ashkboos2024slicegpt} relies on computationally intensive PCA operations and introduces additional parameters. 
Other methods avoid finetuning but introduce new trade-offs: heuristic metrics may compromise performance~\citep{an2024fluctuation, wang2024cfsp}, discrete search spaces become prohibitively large~\citep{shen2024search}, or global loss minimization is needed for mask learning~\citep{gao2024disp}. These gaps highlight the need for structured pruning methods that simultaneously achieve superior performance preservation, hardware-agnostic acceleration, and computational efficiency.

\begin{figure}[!t]
    \centering
    \includegraphics[width=1.0\textwidth]{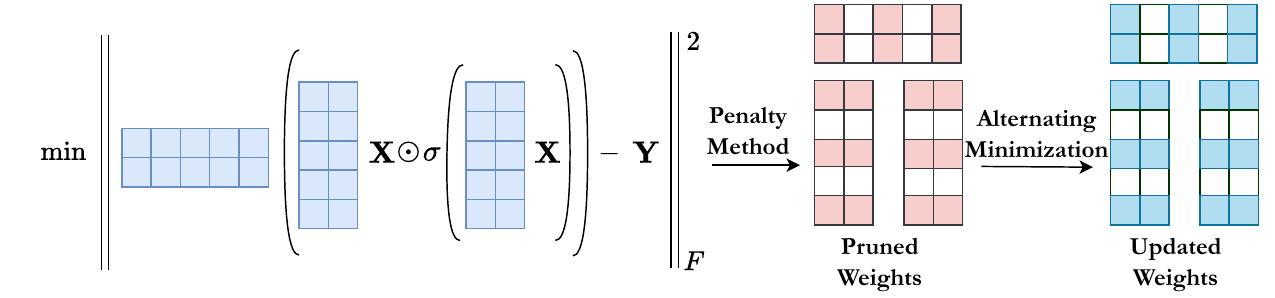}
    \caption{Overview of the SPAP method. \textbf{Left:} The layer-wise pruning problem of an MLP layer. \textbf{Middle:} We propose a penalty method to decide which parts of the weight matrices should be pruned. \textbf{Right:} We develop an alternating minimization algorithm to efficiently update the remaining weights to restore model performance.}
    \label{fig:SPAP_overview}
    \vspace{-0.5cm}
\end{figure}

To address these challenges, we propose SPAP (\underline{S}tructured \underline{P}runing via \underline{A}lternating Optimization and \underline{P}enalty Methods), a novel structured pruning framework grounded in optimization and specifically targeting MLP layers. 
Our approach tackles two core problems: 
(1) identifying optimal neurons to prune with minimal performance impact, and 
(2) efficiently updating remaining parameters for performance recovery. 
For the first challenge, we leverage a structural insight within MLP layers to formulate a mixed-integer optimization problem, rigorously analyze its properties, and derive a theoretical guarantee that relaxation preserves optimality, enabling efficient solutions via a penalty method. 
For the second challenge, we introduce an alternating minimization algorithm that leverages the splittable problem structure by exploiting closed-form updates for one variable and gradient updates for the others, significantly outperforming the vanilla gradient descent approach. Figure~\ref{fig:SPAP_overview} presents an overview of the SPAP method.

Extensive experiments demonstrate SPAP's superior performance across multiple model families (OPT, LLaMA-3/3.1/3.2, and Qwen2.5) on diverse language benchmarks. Our method achieves this while maintaining computational efficiency: pruning the LLaMA-3.1-8B model requires only one hour on a single NVIDIA RTX 4090 GPU (24GB memory) with just 128 calibration samples.
Comprehensive inference profiling reveals that SPAP delivers computational speedups proportional to the achieved sparsity levels, and corresponding reductions in memory footprint. These results collectively validate SPAP's effectiveness and practical utility for real-world deployment scenarios.

\section{Related Work}\label{related_work}
Traditional structured pruning approaches rely on retraining for recovery~\citep{ma2023llm_pruner}, but the computational cost makes this impractical for LLMs, especially in scenarios requiring rapid deployment. This motivates the development of structured pruning methods that preserve performance without retraining.

Recent work explores two primary directions for structured pruning in LLMs. The first focuses on pruning large model components. FinerCut~\citep{zhang2024finercut} employs a greedy algorithm to remove entire attention or MLP blocks with minimal output impact, while ShortGPT~\citep{men2024shortgpt} eliminates decoder layers based on importance metrics. Although effective for inference acceleration, these coarse-grained approaches often incur significant performance drops, necessitating computationally intensive finetuning.

The second direction involves pruning more fine-grained structured components such as rows, columns, or heads in the weight matrices of linear operators, which could better preserve model performance while still being able to provide direct speedups and memory reductions.
SliceGPT~\citep{ashkboos2024slicegpt} introduces rotation-based pruning using PCA-derived orthogonal matrices, but requires storing these matrices for residual connection transformations during inference. 
Furthermore, its activation-dependent strategy demands large calibration datasets and high-precision (64-bit) PCA computations, making it memory-intensive. 
DISP-LLM~\citep{gao2024disp} eliminates structural dependencies by applying distinct selection matrices to weights, obviating residual projection matrices. 
However, it requires training pruning masks via full-network loss minimization, consuming a large amount computational resources (2.4 hours on 2$\times$A100-80GB for 7B models).
Other work~\citep{shen2024search} employs architecture search with ADMM for weight updates, but its iterative search process remains inefficient (5 hours for LLaMA-7B).
SlimGPT~\citep{ling2024slimgpt} and ZipLM~\citep{kurtic2024ziplm} extends the optimal brain surgeon (OBS) framework to structured pruning.
However, the greedy nature of OBS could lead to suboptimal solutions, potentially compromising performance. 
FLAP~\citep{an2024fluctuation} introduces channel stability metrics and bias compensation, but it does not update the remaining weights and thus limits performance recovery. 
While CFSP~\citep{wang2024cfsp} improves upon FLAP by combining inter- and intra-block activation information, it still relies on heuristic metrics and finetuning to recover performance. 
FASP~\citep{hu2025fasp}, proposes a pruning structure that interlinks rows and columns within a decoder layer and develops a prune-update framework for structured pruning. Although FASP achieves relatively good performance with fast pruning speed, the pruning decision is still largely dependent on heuristic metrics.

In summary, while structured pruning offers compelling advantages for LLM deployment, existing methods face three key limitations: (1) dependence on heuristic pruning metrics, (2) inadequate compensation for pruning-induced performance loss, and (3) high computational/memory overhead. Our work addresses these challenges by formulating rigorous optimization models for structured weight pruning, analyzing their mathematical properties, and developing scalable algorithms to enable practical, accurate LLM pruning.

\section{Methodology}
\subsection{Optimization Model for Structured Pruning}\label{problem_formulation}

We formulate the structured pruning problem for MLP layers with gated linear unit (GLU) architecture~\citep{shazeer2020gluvariantsimprovetransformer}, which is widely adopted in state-of-the-art models. Following FASP~\citep{hu2025fasp}, we exploit the structural correspondence between operators by jointly pruning corresponding rows and columns. 
Consider an MLP layer with three linear operators: the up projection $\bm{W}_{up}\in \mathbb{R}^{n\times m}$, gate projection $\bm{W}_{gate}\in \mathbb{R}^{n\times m}$, and down projection $\bm{W}_{down}\in \mathbb{R}^{m\times n}$. Given input $\bm{X} \in \mathbb{R}^{n\times p}$, the forward pass computes:
\begin{equation}
\bm{X}_{\text{output}} = \bm{W}_{down}\left(\bm{W}_{up}\bm{X} \odot \sigma(\bm{W}_{gate}\bm{X})\right),
\end{equation}
where $\sigma(\cdot)$ denotes the swish activation function and $\odot$ represents element-wise multiplication. Through algebraic decomposition, we obtain:
\begin{align}
\bm{X}_{\text{output}} &= \sum_{i=1}^{n} \bm{W}_{down}[:,i]\left(\bm{W}_{up}[i,:]\bm{X} \odot \sigma(\bm{W}_{gate}[i,:]\bm{X})\right),
\end{align}
revealing that pruning the $i$-th column of $\bm{W}_{down}$ permits simultaneous pruning of the $i$-th rows of both $\bm{W}_{up}$ and $\bm{W}_{gate}$ without additional pruning error incurred.

The layer-wise structured pruning problem requires determining which columns to prune in the down projection while optimally updating remaining weights to minimize output distortion. We focus on the down projection to determine which rows/columns should be pruned in the weight matrices as this offers two key advantages: first, it reduces the objective to a tractable least squares formulation; second, it implicitly incorporates the effects of up and gate projections through their activation outputs.
This leads to the following mixed-integer optimization problem with bilinear constraints:
\begin{subequations}
\small
\label{eq:structured_prune_model}
\begin{align}
& \min_{\bm{W}, \bm{s}} \frac{1}{2} \|\bm{WX} - \bm{Y}\|_{F}^2, \label{eq:structured_prune_obj}\\
& \; \text{s.t. } \bm{W} \diag(\bm{s}) = \bm{0}, \label{eq:bilinear}\\
&\qquad \bm{1}^\top \bm{s} = \lambda, \label{eq:simplex} \\
&\qquad \bm{s} \in \{0,1\}^n, \label{eq:binary}
\end{align}
\end{subequations}
where $\bm{W} \in \mathbb{R}^{m \times n}$ represents the pruned weight matrix, $\bm{s} \in \{0,1\}^n$ is a binary pruning indicator variable, $\bm{X} \in \mathbb{R}^{n \times p}$ denotes input activations, $\bm{Y} \in \mathbb{R}^{m \times p}$ is the original output activations, and $\lambda \in \mathbb{N}$ specifies the number of columns to be pruned. The objective function~\eqref{eq:structured_prune_obj} minimizes the Frobenius norm between pruned and original outputs, while constraint~\eqref{eq:bilinear} enforces zero columns for pruned weights, \eqref{eq:simplex} controls sparsity level, and \eqref{eq:binary} ensures binary pruning decisions.

The optimization model presents three significant challenges: its mixed-integer nature from binary constraints, non-convexity introduced by bilinear constraints, and large-scale optimization requirements for LLM parameters. 
We develop solutions to address these challenges through optimization techniques and algorithmic designs in the following.

We first address the mixed-integer challenge of model~\eqref{eq:structured_prune_model} by the following theorem that shows relaxing the integer variables to the continuous space does not affect the optimality of the problem.
\begin{theorem}\label{thm:structured_relaxation}
Consider the following relaxed optimization problem:
\begin{subequations}
\small
\label{eq:structured-relaxed}
\begin{align}
& \min_{\bm{W}, \bm{s}} \frac{1}{2}\|\bm{WX}-\bm{Y}\|_F^2, \\
&\; \text{s.t.} \; \bm{W} \diag(\bm{s}) = 0, \label{eq:relaxed-bilinear} \\
&\quad \quad \bm{1}^\top \bm{s} = \lambda, \label{eq:relaxed-sparsity}\\
&\quad \quad \bm{s} \in [0,1]^n. \label{eq:relaxed-binary}
\end{align}
\end{subequations}
For any optimal solution $(\hat{\bm{W}}, \hat{\bm{s}})$ of (\ref{eq:structured-relaxed}), let $J$ be an arbitrary $\lambda$-subset of the support of $\hat{\bm{s}}$. Then, there exists another optimal solution $(\bm{W}', \bm{s}')$ to (\ref{eq:structured-relaxed}) such that $\bm{s}' \in \{0,1\}^n$ and $\operatorname{supp}(\bm{s}') = J$.
\end{theorem}
Due to page limitations, the proof is provided in Appendix~\ref{proof_relaxation}. 
Theorem~\ref{thm:structured_relaxation} establishes an equivalence relationship: every optimal solution to the relaxed problem~(\ref{eq:structured-relaxed}) contains sufficient information to reconstruct an optimal solution for the original model~(\ref{eq:structured_prune_model}). 
This equivalence justifies our subsequent focus on the relaxed formulation~(\ref{eq:structured-relaxed}). 
The critical challenge emerges from the non-convex bilinear constraint~(\ref{eq:relaxed-bilinear}). 
To address this obstacle, we develop a penalty method in Section~\ref{sec:penalty} that strategically transforms the constrained problem into a sequence of tractable subproblems.

\subsection{A Penalty Method}\label{sec:penalty}
To address the bilinear constraint, we introduce a penalty term, transforming the original problem into:
\small
\begin{align}\label{eq:structured-penalty}
    & \min_{\bm{W}, \bm{s}} \frac{1}{2}\|\bm{WX}-\bm{Y}\|_F^2 + \frac{\rho}{2} \sum_{i=1}^n \bm{s}_i \|\bm{W}[:,i]\|_2^2, \\
    &\; \suchthat \; \bm{1}^\top \bm{s} = \lambda, \\
    &\quad \quad \bm{s} \in [0,1]^n,
\end{align}
where $\rho > 0$ is a penalty parameter controlling the strictness of constraint enforcement.
As $\rho$ increases, the constraints $\bm{s}_i \|\bm{W}[:,i]\|_2^2 = 0$ for all $i \in \{1, \dots, n\}$ are progressively enforced. This ensures that $\bm{W}[:,i] = \bm{0}$ whenever $\bm{s}_i \neq 0$, thereby satisfying $\bm{W} \diag(\bm{s}) = 0$. For sufficiently large $\rho$, the penalized model~(\ref{eq:structured-penalty}) yields the same optimal solutions as the relaxed model~(\ref{eq:structured-relaxed}).

We propose an alternating optimization approach to solve~(\ref{eq:structured-penalty}) approximately. For a given $\rho^{(k)}$, we update $\bm{W}$ and $\bm{s}$ while fixing the other variable:
\small
\begin{numcases}{}
    \bm{s}^{(k+1)} = \argmin\limits_{\bm{s}\in S} \left\{ \frac{1}{2}\|\bm{W}^{(k)}\bm{X}-\bm{Y}\|_F^2 + \frac{\rho^{(k)}}{2} \sum_{i=1}^n \bm{s}_i \|\bm{W}^{(k)}[:,i]\|_2^2 \right\} \label{equ:s-update}, \\
    \bm{W}^{(k+1)} = \argmin\limits_{\bm{W}} \left\{ \frac{1}{2}\|\bm{WX}-\bm{Y}\|_F^2 + \frac{\rho^{(k)}}{2} \sum_{i=1}^n \bm{s}_i^{(k+1)} \|\bm{W}[:,i]\|_2^2 \right\} \label{equ:W-update},
\end{numcases}
where $S := \left\{\bm{s}: \sum_{i=1}^n \bm{s}_i = \lambda, \bm{s} \in [0,1]^n\right\}$ is the feasible set for $\bm{s}$.

%s-update

Let us first investigate the $\bm{s}$-subproblem~(\ref{equ:s-update}). 
Observe that with $\bm{W}^{(k)}$ fixed, the reconstruction error term $\frac{1}{2}\|\bm{WX}-\bm{Y}\|_F^2$ becomes independent of $\bm{s}$, reducing the problem to:
\begin{align}\label{eq:s-subproblem}
    & \min_{\bm{s} \in S} \frac{\rho}{2} \sum_{i=1}^n \bm{s}_i \|\bm{W}^{(k)}[:,i]\|_2^2.
\end{align}
This could be naively solved by setting $\lambda$ entries in $\bm{s}^{(k+1)}$ corresponding to the $\lambda$ columns with smallest norms to 1 and the remaining entries to 0. 
However, this approach is myopic as it only considers the magnitude of the current $\bm{W}^{(k)}$ while neglecting our ultimate objective of minimizing the reconstruction error.
Inspired by Wanda~\citep{sun2023simple}, we incorporate the impact of reconstruction error by jointly considering both the column norms of $\bm{W}^{(k)}$ and the column-wise Wanda score. 
We define the following composite metric:
\begin{equation}
    \text{score}(\bm{s}_j) := t \left\|\bm{W}^{(k)}[:,j]\right\|_2^2 + (1-t) \left\|\bm{W}^{(k)}[:,j] \right\|_1 \cdot \left\|\bm{X}[:, j]\right\|_2,
\end{equation}
where $t \in (0,1)$ is a hyperparameter balancing these two terms. 
The $\lambda$ entries of $\bm{s}$ corresponding to the smallest scores are then set to 1, with the remaining entries set to 0.
To promote smoother optimization while encouraging exploration and leveraging the continuous relaxation of $\bm{s}$, we employ a soft update mechanism:
\begin{equation}
    \bm{s}^{(k+1)} = \alpha \bm{s}^{(k)} + (1-\alpha) \bm{s}^{(k+1)}_{\text{new}},
\end{equation}
where $\alpha \in (0,1)$ controls the update aggressiveness.

%%%%%%%%%%%%%%%%%see if we need to put this into the appendix
The $\bm{W}$-subproblem~(\ref{equ:W-update}) is a generalized ridge regression problem with column-specific regularization governed by $\bm{s}$.
It admits the following closed-form solution (please see Appendix~\ref{W-derivation} for mathematical derivations):
$$
\bm{W}^{(k+1)} = \bm{Y} \bm{X}^\top (\bm{X} \bm{X}^\top + \rho^{(k)} \diag(\bm{s}^{(k+1)}))^{-1}.
$$
In case the matrix $\bm{X} \bm{X}^\top + \rho^{(k)} \diag(\bm{s}^{(k+1)})$ is only positive semi-definite, a small diagonal perturbation $\delta \bm{I}$ ($\delta > 0$) may be added for numerical stability.

After obtaining the updated variables $\bm{s}^{(k+1)}$ and $\bm{W}^{(k+1)}$, we increase the penalty parameter $\rho$ according to the following scheme to progressively enforce the sparsity constraints:
$$
\rho^{(k+1)} = \tau \rho^{(k)},
$$
where $\tau > 1$ controls the rate of increase. 
This adaptive scheme ensures stronger constraint satisfaction as the optimization progresses.

After a given $K$ iterations, we do a final hard update step of $\bm{s}$ and update $\bm{W}$ accordingly to recover a feasible solution of model~\eqref{eq:structured_prune_model}.
The described penalty method for solving model~(\ref{eq:structured-relaxed}) is summarized in Algorithm~\ref{alg:penalty}.
\begin{algorithm}[t]
    \small
    \caption{A penalty method for model~(\ref{eq:structured-relaxed}) }
    \label{alg:penalty}
    \begin{algorithmic}
    \STATE \textbf{Inputs:} \\
    \quad original weight $\bm{W} \in \mathbb{R}^{m \times n}$, \\
    \quad input activation $\bm{X} \in \mathbb{R}^{n \times p}$, \\
    \quad output activation of the original network $\bm{Y} \in \mathbb{R}^{m \times p}$, \\
    \quad \# of columns to be pruned $\lambda \in \mathbb{N}$, \\
    \quad \# of iterations $K$, score parameter $t$, soft update parameter $\alpha$, increase parameter $\tau$ \\
    \STATE \textbf{Output:} \\
    \quad pruned and updated weight $\bm{W}^* \in \mathbb{R}^{m \times (n-\lambda)}$ \\

    Initialize $\bm{s}^{(0)}$, $\bm{W}^{(0)}$ and $\rho^{(0)}$
    \FOR{$k \leftarrow$ 0 to K-1}
        \FOR{$j \leftarrow$ 1 to n}
            \STATE
            $\text{score}\left(\bm{s}_j^{(k)}\right) \leftarrow t \left\|\bm{W}^{(k)}[:,j]\right\|_2^2 + (1-t) \left\|\bm{W}^{(k)}[:,j] \right\|_1 \cdot \left\|\bm{X}[:, j]\right\|_2$, \COMMENT{\textcolor{teal}{Compute scores}}
        \ENDFOR
        \FOR{$l \leftarrow$ 1 to n}
            \IF{$\text{score}\left(\bm{s}_j^{(k)}\right) \leq \mathrm{Top}_{n-\lambda}\left(\text{score}\left(\bm{s}_j^{(k-1)}\right)\right)$}
                \STATE $\bm{s}_l^{(k+\frac12)} \leftarrow$ 1 \COMMENT{\textcolor{teal}{Compute the new $\bm{s}$}}
            \ENDIF
        \ENDFOR
        \STATE$\bm{s}^{(k+1)} = \alpha \bm{s}^{(k)} + (1-\alpha) \bm{s}^{(k+\frac12)}$ \COMMENT{\textcolor{teal}{Soft update $\bm{s}$}}\\
        \STATE$\bm{W}^{(k+1)} = \bm{Y} \bm{X}^\top (\bm{X} \bm{X}^\top + \rho^{(k)} \diag(\bm{s}^{(k+1)}))^{-1}$ \COMMENT{\textcolor{teal}{Update $\bm{W}$}} \\
        \STATE$\rho^{(k+1)} = \tau \rho^{(k)}$ \COMMENT{\textcolor{teal}{Increase the penalty parameter $\rho$}}
    \ENDFOR
    \STATE Compute $\text{score}\left(\bm{s}^{(K)}\right)$ and conduct hard update to get $\bm{s}^*$ \\
    \STATE Updated $\bm{W}^*$ according to $\bm{s}^*$
    \STATE \textbf{return} $\bm{W}^*$
    \end{algorithmic}
\end{algorithm}
% \vspace{-0.5cm}

\subsection{Updating the Up and Gate Projections}
Having established our penalty method for pruning and updating the down projections in MLP blocks, we now address the remaining components: the up and gate projections. While these weight matrices are pruned following the correspondence rules, they remain unupdated, missing opportunities to further reduce the pruning error. To bridge this gap, we develop an efficient alternating minimization algorithm that jointly optimizes all three projection matrices in the pruned architecture. This approach systematically reduces pruning error while maintaining computational efficiency.

Let $\bm{W}_{up} \in \mathbb{R}^{(n-\lambda)\times m}$, $\bm{W}_{gate} \in \mathbb{R}^{(n-\lambda)\times m}$, and $\bm{W}_{down} \in \mathbb{R}^{m\times (n-\lambda)}$ denote the weight matrices of the pruned up, gate, and down projections, respectively. Given input activations $\bm{X} \in \mathbb{R}^{m\times p}$ and original output activations $\bm{Y} \in \mathbb{R}^{m\times p}$ of the MLP block, we formulate the weight update problem as the following unconstrained optimization problem:
\begin{align}
    \label{eq:update-up-gate}
    \min_{\bm{W}_{up}, \bm{W}_{gate}, \bm{W}_{down}}  
    \left\|
        \bm{W}_{down} \left( \bm{W}_{up}\bm{X} \odot \sigma(\bm{W}_{gate}\bm{X}) \right) - \bm{Y}
    \right\|_2^2.
\end{align}
Although unconstrained, Problem~\eqref{eq:update-up-gate} presents computational challenges due to the complex interactions among decision variables and the non-convexity introduced by the Swish activation function $\sigma(\cdot)$. While gradient descent offers a straightforward approach for optimizing these variables, we observe that when $\bm{W}_{up}$ and $\bm{W}_{gate}$ are fixed, the problem reduces to a least squares formulation with the closed-form solution:
\begin{equation}
    \bm{W}_{down} = \bm{Y}\bm{Z}^\top \left(\bm{Z} \bm{Z}^\top\right)^{-1},
\end{equation}
where $\bm{Z} = \bm{W}_{up}\bm{X} \odot \sigma(\bm{W}_{gate}\bm{X})$. This observation motivates our more efficient alternating minimization approach, which combines gradient descent steps for $\bm{W}_{up}$ and $\bm{W}_{gate}$ with an optimal update for $\bm{W}_{down}$:
\begin{subequations}
\small
\begin{align}\label{algo:SPAP-alternating}
    \bm{W}_{up}^{(k+1)} &= \bm{W}_{up}^{(k)} - \eta \frac{\partial f}{\partial \bm{W}_{up}}
    \left(\bm{W}_{up}^{(k)},\bm{W}_{gate}^{(k)},\bm{W}_{down}^{(k)}\right), \\
    \bm{W}_{gate}^{(k+1)} &= \bm{W}_{gate}^{(k)} - \eta \frac{\partial f}{\partial \bm{W}_{gate}}\left(\bm{W}_{up}^{(k)},\bm{W}_{gate}^{(k)},\bm{W}_{down}^{(k)}\right), \\
    \bm{Z}^{(k+1)} &= \bm{W}_{gate}^{(k+1)}\bm{X} \odot \sigma\left(\bm{W}_{up}^{(k+1)}\bm{X}\right), \\
    \bm{W}_{down}^{(k+1)} &= \bm{Y}\left(\bm{Z}^{(k+1)}\right)^\top \left(\bm{Z}^{(k+1)} \left(\bm{Z}^{(k+1)}\right)^\top\right)^{-1}.
\end{align}
\end{subequations}
Notably, with $\bm{W}_{down}$ and $\bm{W}_{gate}$ fixed, the problem reduces to a least squares formulation for $\bm{W}_{up}$. However, solving this optimally would require vectorization and Kronecker products, which leads to prohibitive memory requirements when pruning LLMs. Consequently, we maintain memory-efficient gradient updates for $\bm{W}_{up}$. In our implementation, we employ the Adam optimizer~\citep{kingma2014adam} to automatically adapt the learning rate $\eta$ during optimization. Section~\ref{chap4:SPAP_ablation} presents a comprehensive comparison between our alternating optimization approach and the vanilla gradient descent method, showcasing the consistently better performance of the proposed method.

% perplexity
\begin{figure}[!ht]
    \centering
    \subfigure[Perplexity-vs-sparsity on LLaMA-3.2-1B.]{
        \includegraphics[width=0.45\textwidth]{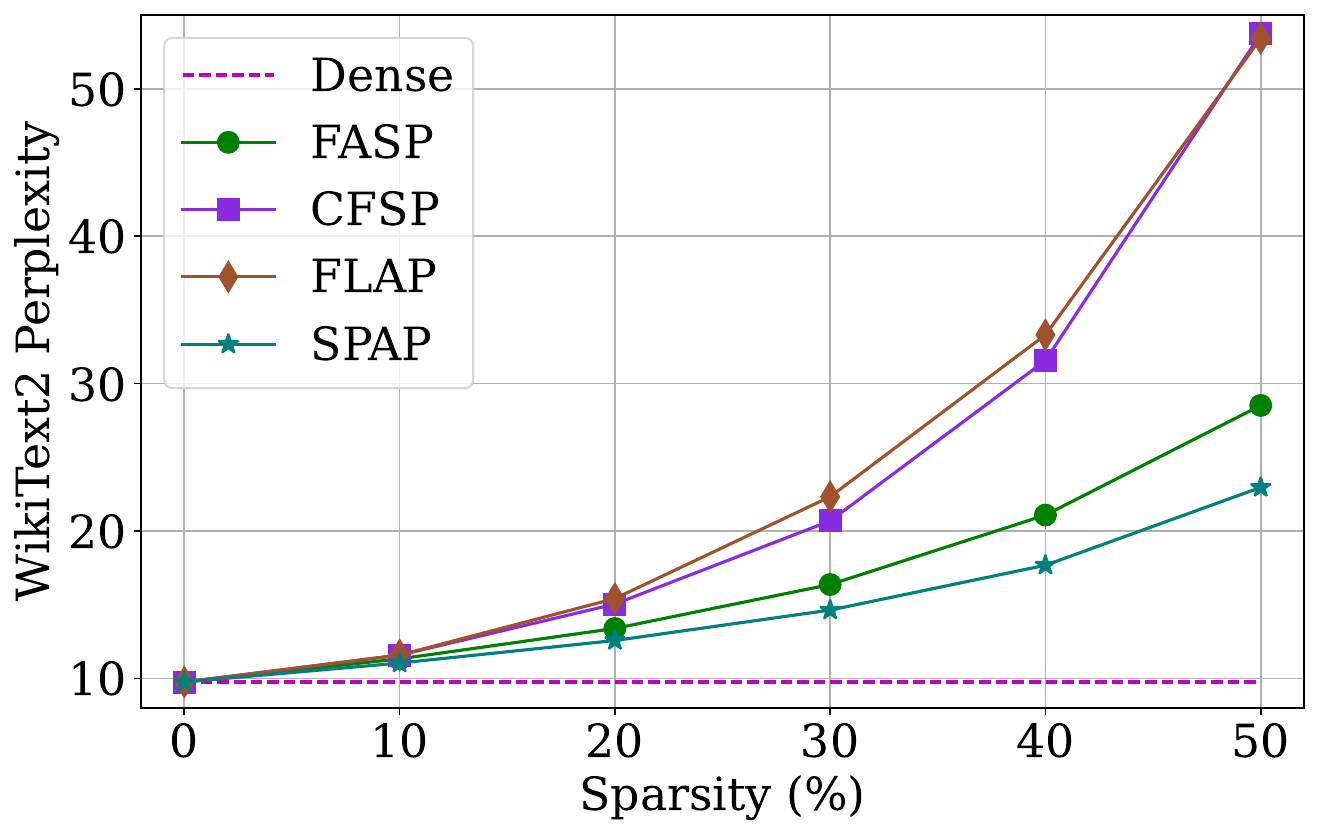}
        \label{fig:SPAP-llama-3.2-1B}
    }
    \hspace{0.5cm}
    \subfigure[Perplexity-vs-sparsity on Qwen2.5-0.5B.]{
        \includegraphics[width=0.45\textwidth]{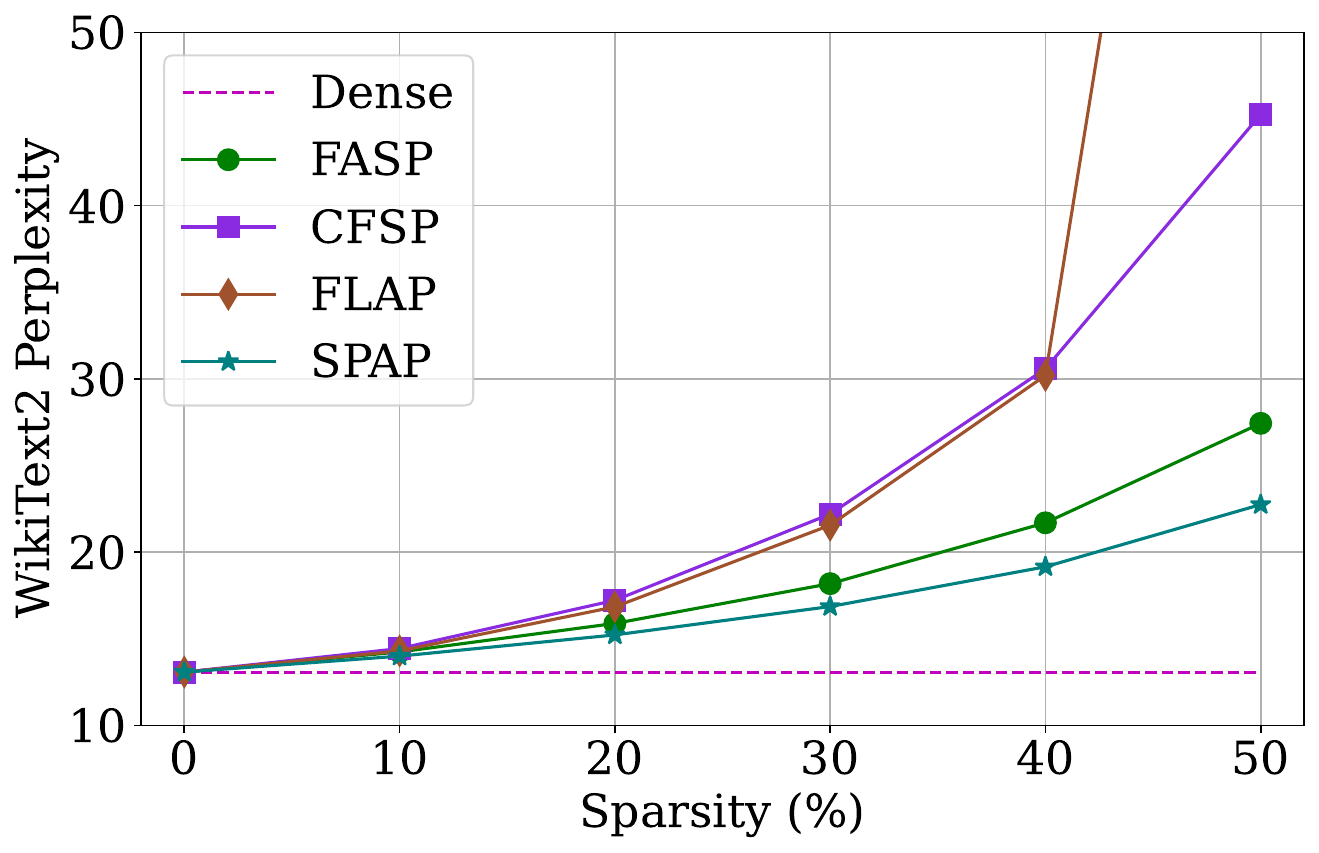}
        \label{fig:SPAP-qwen2.5-0.5B}
    }
    \caption{Perplexity results of pruned LLaMA-3.2-1B and Qwen2.5-0.5B models under various sparsity. SPAP achieves significantly lower ppl in all settings than the baseline methods.}
    \label{fig:SPAP-various-sparsity}
    \vspace{-0.2cm}
\end{figure}
\section{Experiments}

\subsection{Experimental Setup}
\noindent\textbf{Models and Baselines.} We compare SPAP against four competitive pruning methods: CFSP~\citep{wang2024cfsp}, FLAP~\citep{an2024fluctuation}, SliceGPT~\citep{ashkboos2024slicegpt}, and FASP~\citep{hu2025fasp}. 
Our evaluation spans five prominent model families: LLaMA-3/3.1/3.2 series~\citep{meta2023llama3, grattafiori2024llama3-1}, Qwen2.5 series~\citep{qwen2025qwen25technicalreport}, and OPT series~\citep{zhang2022opt} with sizes ranging from 125M to 8B. 
Due to lack of support from the official implementation, SliceGPT comparisons are limited to OPT models, while FLAP and CFSP evaluations exclude OPT models. 
All models were sourced from HuggingFace~\citep{wolf2019huggingface}.
The pruning via CFSP, FLAP, FASP and SPAP are only on the MLP blocks and we scale up the the actual pruning ratio by the inverse of the proportion of parameters in the MLP blocks so that all the sparsity figures represent overall sparsity of the whole model.
The SliceGPT method is applied on both attention and MLP blocks as the rotation matrices couples the pruning of the two blocks and we donot account for the additional parameters introduced in the orthogonal matrices for modified residual connections so that the actual model sparsity of SliceGPT is lower (e.g., 20\% pruning via SliceGPT yields only 10.1\% actual parameter reduction in OPT-125M). 

\noindent\textbf{Datasets and benchmarks.} Following prior works~\citep{sun2023simple,frantar2023sparsegpt}, we employ 128 randomly selected calibration samples from the WikiText2~\citep{merity2016pointer} dataset for SPAP and the baseline methods. 
Full WikiText perplexity evaluation of the pruned models are then conducted to assess the pruning performance.
For reasoning tasks, we evaluate zero-shot accuracy on six benchmarks: ARC-easy, ARC-challenge~\citep{clark2018think}, OpenbookQA~\citep{mihaylov2018can}, PIQA~\citep{bisk2020piqa}, WinoGrande~\citep{sakaguchi2021winogrande}, and RTE~\citep{wang2018glue}, using the LM Harness library~\citep{gao2021framework}.

\noindent\textbf{Implementation Details.} All experiments are conducted on a single NVIDIA RTX 4090 GPU with 24GB memory. We implement SPAP using the PyTorch~\citep{paszke2019pytorch} library. 
For baseline comparisons, we implement FASP based on the specifications in their paper; employ the official implementations of CFSP and SliceGPT; and derive FLAP results using CFSP's implementation, as the original FLAP code lacks support for LLaMA-3.1/3.2 and Qwen2.5 models. 
SPAP is able to prune LLaMA-3.1-8B on a single 4090 GPU in approximately 1 hour.

\subsection{Main Results}

\begin{center}
    \begin{minipage}[t]{0.47\textwidth}
        \centering
        \small
        \setlength{\tabcolsep}{4.5pt}
        \renewcommand{\arraystretch}{0.85}
        \begin{adjustbox}{max width=\linewidth,valign=t}
        \begin{tabular}{l | c | c c c c}
        \toprule 
        \multirow{2}{*}{Method} & \multirow{2}{*}{Sparsity} & \multicolumn{4}{c}{OPT} \\
        \cmidrule(lr){3-6}
        & & 125M & 1.3B & 2.7B & 6.7B \\
        \midrule
        Dense & 0$\%$ & 27.66 & 14.63 & 12.47 & 10.86 \\
        \cmidrule(lr){1-6}
        SliceGPT & 10$\%$ & 29.32 & 15.15 & 14.10 & 11.01 \\
        FASP & 10$\%$ & 28.35 & 14.85 & 12.51 & 10.86 \\
        \cellcolor{yellow!25}SPAP & \cellcolor{yellow!25}10$\%$ & \cellcolor{yellow!25}\textbf{28.09} & \cellcolor{yellow!25}\textbf{14.77} & \cellcolor{yellow!25}\textbf{12.41} & \cellcolor{yellow!25}\textbf{10.83} \\
        \cmidrule(lr){1-6}
        SliceGPT & 20$\%$ & 34.53 & 16.59 & 16.81 & 11.62 \\
        FASP & 20$\%$ & 29.93 & 16.04 & 13.79 & 11.59 \\
        \cellcolor{yellow!25}SPAP & \cellcolor{yellow!25}20$\%$ & \cellcolor{yellow!25}\textbf{29.33} & \cellcolor{yellow!25}\textbf{15.66} & \cellcolor{yellow!25}\textbf{13.28} & \cellcolor{yellow!25}\textbf{11.40} \\
        \cmidrule(lr){1-6}
        SliceGPT & 30$\%$ & 44.63 & 19.60 & 24.12 & 14.21 \\
        FASP & 30$\%$ & 34.06 & 18.65 & 17.01 & 13.31 \\
        \cellcolor{yellow!25}SPAP & \cellcolor{yellow!25}30$\%$ & \cellcolor{yellow!25}\textbf{32.55} & \cellcolor{yellow!25}\textbf{17.90} & \cellcolor{yellow!25}\textbf{15.39} & \cellcolor{yellow!25}\textbf{12.90} \\
        \bottomrule
        \end{tabular}
        \end{adjustbox}
        \captionof{table}{WikiText perplexity ($\downarrow$) of pruned OPT models under various sparsity.}
        \label{tab:SPAP_ppl_results_opt}
    \end{minipage}
    \hfill
    \begin{minipage}[t]{0.475\textwidth}
        \vspace{-\baselineskip} % 补偿标题的垂直空间
        \textbf{Perplexity Results.} 
        Tables~\ref{tab:SPAP_ppl_results_opt} and \ref{tab:SPAP_ppl_results_combined} demonstrate SPAP's consistent superiority across all OPT, LLaMA-3/3.1/3.2 and Qwen2.5 model families under 10\%, 20\% and 30\% structured sparsity levels. 
        SPAP maintains lower perplexity than all baselines at every sparsity level, with the advantage becoming more pronounced for more challenging pruning tasks where higher sparsity levels and smaller dense model are considered. 
        For Qwen2.5-1.5B and 7B models, CFSP and FLAP fail to produce reasonable results, with perplexity equals to NaN. Figure~\ref{fig:SPAP-various-sparsity} illustrates the perplexity trend as sparsity increases from 0\% to 50\% for LLaMA-3.2-1B and Qwen2.5-0.5B models.
        We observe a clear advantage of SPAP compared with baseline methods, especially against CFSP and FLAP, across a wide range of sparsity levels.
    \end{minipage}
\end{center}
%%%%%%%%%%%%%%

% llama qwen
\begin{table*}[ht!]
    \centering
    \small
    \setlength{\tabcolsep}{4.5pt}
    \renewcommand{\arraystretch}{1.05}
    \resizebox{0.78\textwidth}{!}{
    \begin{tabular}{l | c | c c c c | c c c c}
    \toprule 
    \multirow{2}{*}{Method} & \multirow{2}{*}{Sparsity} & \multicolumn{4}{c|}{LLaMA-3.x} & \multicolumn{4}{c}{Qwen2.5} \\
    \cmidrule(lr){3-6} \cmidrule(lr){7-10}
    & & 0-8B & 1-8B & 2-1B & 2-3B & 0.5B & 1.5B & 3B & 7B \\
    \midrule
    Dense & 0$\%$ & 6.14 & 6.24 & 9.76 & 7.81 & 13.08 & 9.28 & 8.03 & 6.85 \\
    \cmidrule(lr){1-10}
    CFSP & 10$\%$ & 7.17 & 7.20 & 11.58 & 9.08 & 14.44 & - & 8.96 & - \\
    FASP & 10$\%$ & 7.11 & 7.11 & 11.33 & 8.86 & 14.24 & 10.11 & 9.04 & 7.30 \\
    FLAP & 10$\%$ & 7.20 & 7.27 & 11.59 & 9.10 & 14.30 & - & \textbf{8.51} & - \\
    \cellcolor{yellow!25}SPAP & \cellcolor{yellow!25}10$\%$ & \cellcolor{yellow!25}\textbf{7.03} & \cellcolor{yellow!25}\textbf{7.07} & \cellcolor{yellow!25}\textbf{11.05} & \cellcolor{yellow!25}\textbf{8.69} & \cellcolor{yellow!25}\textbf{13.99} & \cellcolor{yellow!25}\textbf{10.00} & \cellcolor{yellow!25}8.91 & \cellcolor{yellow!25}\textbf{7.25} \\
    \cmidrule(lr){1-10}
    CFSP & 20$\%$ & 8.67 & 8.71 & 15.05 & 11.47 & 17.22 & - & 10.72 & - \\
    FASP & 20$\%$ & 8.34 & 8.33 & 13.39 & 10.69 & 15.89 & 11.41 & 10.16 & 7.96 \\
    FLAP & 20$\%$ & 8.91 & 8.89 & 15.42 & 11.77 & 16.84 & - & 9.91 & - \\
    \cellcolor{yellow!25}SPAP & \cellcolor{yellow!25}20$\%$ & \cellcolor{yellow!25}\textbf{8.17} & \cellcolor{yellow!25}\textbf{8.12} & \cellcolor{yellow!25}\textbf{12.58} & \cellcolor{yellow!25}\textbf{10.27} & \cellcolor{yellow!25}\textbf{15.22} & \cellcolor{yellow!25}\textbf{11.07} & \cellcolor{yellow!25}\textbf{9.85} & \cellcolor{yellow!25}\textbf{7.81} \\
    \cmidrule(lr){1-10}
    CFSP & 30$\%$ & 11.16 & 11.12 & 20.69 & 15.62 & 22.20 & - & 13.59 & - \\
    FASP & 30$\%$ & 10.44 & 10.37 & 16.37 & 13.92 & 18.18 & 12.97 & 11.51 & 8.81 \\
    FLAP & 30$\%$ & 11.77 & 11.59 & 22.33 & 15.98 & 21.56 & - & 12.86 & - \\
    \cellcolor{yellow!25}SPAP & \cellcolor{yellow!25}30$\%$ & \cellcolor{yellow!25}\textbf{9.90} & \cellcolor{yellow!25}\textbf{9.88} & \cellcolor{yellow!25}\textbf{14.64} & \cellcolor{yellow!25}\textbf{12.53} & \cellcolor{yellow!25}\textbf{16.86} & \cellcolor{yellow!25}\textbf{12.34} & \cellcolor{yellow!25}\textbf{11.21} & \cellcolor{yellow!25}\textbf{8.58} \\
    \bottomrule
    \end{tabular}
    }
    \caption{WikiText perplexity ($\downarrow$) of pruned LLaMA-3.x and Qwen2.5 models under various sparsity. SPAP outperforms baseline methods.}
    \label{tab:SPAP_ppl_results_combined}
\end{table*}

% zeroshot
\begin{table*}[ht!]
    \centering
    \small
    \setlength{\tabcolsep}{5.0pt}
    \renewcommand{\arraystretch}{1.0}
    \resizebox{0.9\textwidth}{!}{
    % \begin{tabular}{l l c c c c c c c}
    \begin{tabular}{l | l | c c c c c c | c}
    \toprule 
    Model & Method & ARC-e  & ARC-c  & PIQA   & OBQA   & WinoGrande  & RTE   & Mean \\
    \midrule
    \multirow{5}{*}{LLaMA-3.1-8B} 
    & Dense  & 81.14  & 53.24  & 81.12  & 44.80  & 73.80  & 71.12  & 67.54 \\
    \cmidrule(lr){2-9}
    & CFSP   & 74.96  & 47.27  & 78.56  & 42.20  & 73.48  & 63.90  & 63.40  \\
    & FLAP   & 67.76  & 40.61  & 76.77  & 41.40  & 69.77  & 53.07  & 58.23  \\
    & FASP   & 75.00  & 47.18  & 79.11  & 42.40  & 71.27  & 58.84  & 62.30  \\
    & \cellcolor{yellow!25}SPAP & \cellcolor{yellow!25}75.88 & \cellcolor{yellow!25}47.78 & \cellcolor{yellow!25}78.40 & \cellcolor{yellow!25}43.40 & \cellcolor{yellow!25}73.01 & \cellcolor{yellow!25}70.76 & \cellcolor{yellow!25}\textbf{64.89} \\
    \cmidrule(lr){1-9}
    \multirow{5}{*}{Qwen2.5-3B} 
    & Dense  & 73.02  & 47.10  & 78.40  & 42.60  & 67.96  & 75.09  & 64.02 \\
    \cmidrule(lr){2-9}
    & CFSP   & 68.81  & 43.09  & 75.52  & 42.40  & 64.96  & 61.73  & 59.42  \\
    & FLAP   & 73.69  & 45.22  & 74.81  & 42.00  & 67.56  & 68.59  & 61.98  \\
    & FASP   & 70.33  & 44.11  & 75.46  & 42.40  & 65.19  & 75.45  & 62.16  \\
    & \cellcolor{yellow!25}SPAP & \cellcolor{yellow!25}72.52 & \cellcolor{yellow!25}45.22 & \cellcolor{yellow!25}75.46 & \cellcolor{yellow!25}42.60 & \cellcolor{yellow!25}66.61 & \cellcolor{yellow!25}82.67 & \cellcolor{yellow!25}\textbf{64.18} \\
    \bottomrule
    \end{tabular}
    }
    \caption{Zero-shot results (accuracy, $\uparrow$) of the pruned models under 10\% sparsity. SPAP outperforms baseline methods for both LLaMA-3.1-8B and Qwen2.5-3B models.}
    \label{tab:SPAP_zero_shot_combined}
\end{table*}

\noindent\textbf{Zero-shot Performance.} 
As demonstrated in Table~\ref{tab:SPAP_zero_shot_combined}, SPAP maintains strong reasoning capabilities across diverse tasks under 10\% structured sparsity. 
On LLaMA-3.1-8B, SPAP achieves a mean accuracy of \textbf{64.89\%}, outperforming CFSP (+1.49\%), FLAP (+6.66\%), and FASP (+2.59\%).
For Qwen2.5-3B, SPAP reaches \textbf{64.18\%} mean accuracy, surpassing all baselines by 2.02\%--4.76\%. 
The consistent improvements across both model families demonstrate SPAP's robustness in maintaining model capabilities during compression.

\noindent\textbf{Inference Profiling}.
We profile the CUDA execution time and peak memory usage during inference of pruned LLaMA-3.1-8B and Qwen2.5-7B models on a single NVIDIA RTX 4090 GPU. For each test, 1024 tokens are generated.
As shown in Table~\ref{tab:SPAP-profile}, SPAP achieves linear speedups and memory reductios proportional to the sparsity level. Specifically, at 30\% sparsity, LLaMA-3.1-8B demonstrates a 1.28$\times$ speedup with 26\% memory reduction, while Qwen2.5-7B shows a 1.29$\times$ speedup with 23\% memory savings. The improvements scale consistently across sparsity levels, confirming SPAP's hardware efficiency. Notably, even moderate sparsity (10-20\%) yields notable gains without significantly compromising model quality, as evidiented in the mentioned perplexity and zero-shot results.

% profile
\begin{table*}[ht!]
    \centering
    \small
    \setlength{\tabcolsep}{6.5pt}
    \renewcommand{\arraystretch}{1.0}
    \resizebox{0.78\textwidth}{!}{
    \begin{tabular}{l | l | c c | c c}
    \toprule 
    \multirow{2}{*}{Model} & \multirow{2}{*}{Sparsity} & CUDA & \multirow{2}{*}{Speedup} & Peak  & Memory  \\
     & & Time (s) & & Memory (GB) & Reduction \\
    \midrule
    \multirow{4}{*}{LLaMA-3.1-8B} 
    & Dense & 19.03 & 1.00 & 15.09 & 1.00 \\
    \cmidrule(lr){2-6}
    & 10\% & 17.80 & 1.07 & 13.79 & 0.91 \\
    & 20\% & 16.29 & 1.17 & 12.49 & 0.83 \\
    & 30\% & 14.85 & 1.28 & 11.19 & 0.74 \\
    \midrule
    \multirow{4}{*}{Qwen2.5-7B} 
    & Dense & 17.75 & 1.00 & 16.04 & 1.00 \\
    \cmidrule(lr){2-6}
    & 10\% & 16.73 & 1.06 & 14.78 & 0.92 \\
    & 20\% & 15.26 & 1.16 & 13.58 & 0.85 \\
    & 30\% & 13.71 & 1.29 & 12.43 & 0.77 \\
    \bottomrule
    \end{tabular}
    }
    \caption{Inference latency and memory footprint of pruned LLaMA-3.1-8B and Qwen2.5-7B models at varying sparsity levels. Speedup and memory reduction are normalized to dense baselines.}
    \label{tab:SPAP-profile}
\end{table*}

% ablation
\begin{table*}[ht!]
    \centering
    \small
    \setlength{\tabcolsep}{6.5pt}
    \renewcommand{\arraystretch}{1.0}
    \resizebox{0.8\textwidth}{!}{
    \begin{tabular}{l|l|ccccc}
        \toprule
        \multirow{2}{*}{Model} & \multirow{2}{*}{Method} & \multicolumn{5}{c}{Sparsity} \\
        \cmidrule(lr){3-7}
            & & 10\% & 20\% & 30\% & 40\% & 50\% \\
        \midrule
        \multirow{6}{*}{Qwen2.5-0.5B}
        &FASP & 14.24 & 15.89 & 18.18 & 21.69 & 27.44 \\
        & CFSP & 14.44 & 17.22 & 22.20 & 30.58 & 45.26 \\
        & FLAP & 14.30 & 16.84 & 21.56 & 30.20 & 107.10 \\
        & \cellcolor{yellow!25}SPAP w/o update & \cellcolor{yellow!25}14.17 & \cellcolor{yellow!25}15.58 & \cellcolor{yellow!25}17.69 & \cellcolor{yellow!25}20.82 & \cellcolor{yellow!25}26.23 \\
        & \cellcolor{yellow!25}SPAP w. GD & \cellcolor{yellow!25}14.15 & \cellcolor{yellow!25}15.53 & \cellcolor{yellow!25}17.50 & \cellcolor{yellow!25}20.23 & \cellcolor{yellow!25}24.92 \\
        & \cellcolor{yellow!25}\textbf{SPAP} & \cellcolor{yellow!25}\textbf{14.00} & \cellcolor{yellow!25}\textbf{15.22} & \cellcolor{yellow!25}\textbf{16.86} & \cellcolor{yellow!25}\textbf{19.16} & \cellcolor{yellow!25}\textbf{22.75} \\
        \midrule
        \multirow{6}{*}{LLaMA-3.2-1B}
        & FASP & 11.33 & 13.39 & 16.37 & 21.08 & 28.52 \\
        & CFSP & 11.58 & 15.05 & 20.69 & 31.57 & 53.71 \\
        & FLAP & 11.59 & 15.42 & 22.33 & 33.31 & 53.41 \\
        & \cellcolor{yellow!25}SPAP w/o update & \cellcolor{yellow!25}11.16 & \cellcolor{yellow!25}12.91 & \cellcolor{yellow!25}15.47 & \cellcolor{yellow!25}19.51 & \cellcolor{yellow!25}26.21 \\
        & \cellcolor{yellow!25}SPAP w. GD & \cellcolor{yellow!25}11.15 & \cellcolor{yellow!25}12.89 & \cellcolor{yellow!25}15.40 & \cellcolor{yellow!25}19.47 & \cellcolor{yellow!25}26.08 \\
        & \cellcolor{yellow!25}\textbf{SPAP} & \cellcolor{yellow!25}\textbf{11.05} & \cellcolor{yellow!25}\textbf{12.58} & \cellcolor{yellow!25}\textbf{14.64} & \cellcolor{yellow!25}\textbf{17.68} & \cellcolor{yellow!25}\textbf{22.96} \\
        \bottomrule
    \end{tabular}
    }
    \caption{Perplexity results ($\downarrow$) of baseline methods and ablated varients on Qwen2.5-0.5B and LLaMA-3.2-1B models across sparsity levels.}
    \label{tab:SPAP-ablation}
\end{table*}

\subsection{Ablation Study}\label{chap4:SPAP_ablation}

We conduct comprehensive ablation studies to systematically evaluate the effectiveness of both the proposed penalty method and the alternating minimization algorithm. 
Table~\ref{tab:SPAP-ablation} presents the perplexity results of pruned LLaMA-3.2-1B and Qwen2.5-0.5B models, comparing our approach with three baseline methods (FASP, CFSP, and FLAP) as well as two ablated variants of our method. 
The first variant, denoted as \textit{SPAP w/o update}, applies only the penalty method without executing the alternating minimization algorithm to update the operators. 
The second variant, \textit{SPAP w. GD}, implements the penalty method followed by gradient descent optimization using the Adam optimizer to solve model~(\ref{eq:update-up-gate}). 
Both SPAP w. GD and our full SPAP method perform 20 iterations of updates for fair comparison. 

Several key observations emerge from the experimental results. 
First, even the basic SPAP w/o update variant demonstrates superior performance compared to all baseline methods, achieving consistently lower perplexity scores across all sparsity levels. 
This improvement is particularly notable when compared to FASP, as it validates that our penalty method can indeed discover better structured pruning masks than those obtained from the columnwise Wanda score approach. 

Furthermore, the comparison between SPAP w. GD and our full SPAP method reveals a significant performance gap, with SPAP achieving substantially better perplexity results. This empirical evidence strongly supports the effectiveness of our proposed alternating minimization algorithm over conventional gradient descent optimization. The progressive improvement from SPAP w/o update to SPAP w. GD and finally to our complete SPAP method demonstrates the complementary benefits of both the penalty formulation and the specialized optimization strategy.

\section{Conclusion}
We propose SPAP, an optimization-driven structured pruning framework for LLMs that formulates pruning as a mixed-integer problem, solves it via penalty methods and alternating minimization, and achieves linear speedups (1.29$\times$ at 30\% sparsity) and memory reductions with minimal accuracy loss. 
Comprehensive experiments on OPT, LLaMA-3/3.1/3.2, and Qwen2.5 model families demonstrate SPAP's superiority over state-of-the-art baselines in both perplexity and zero-shot task accuracy. 
Our work establishes a principled optimization approach for scalable pruning, which opens new possibilities for structured pruning for LLMs.

\clearpage
{
\bibliography{ref}
\bibliographystyle{apalike}
}

%%%%%%%%%%%%%%%%%%%%%%%%%%%%%%%%%%%%%%%%%%%%%%%%%%%%%%%%%%%%
\clearpage
\appendix

\section{Proof of Theorem~\ref{thm:structured_relaxation}}\label{proof_relaxation}

We provide a constructive proof of Theorem~\ref{thm:structured_relaxation}, establishing that the original mixed-integer optimization model~\eqref{eq:structured_prune_model} and its relaxed counterpart~\eqref{eq:structured-relaxed} share identical optimal solutions. Specifically, any optimal solution to the continuous relaxation can be directly transformed into an optimal binary solution for the original integer-constrained problem by construction. This equivalence demonstrates that relaxing the discrete variables in model~\eqref{eq:structured_prune_model} preserves the global optimality while enabling more efficient numerical treatment.

\begin{proof}
Let $(\hat{\bm{W}}, \hat{\bm{s}})$ be an optimal solution to the relaxed problem~(\ref{eq:structured-relaxed}). We analyze the constraints and construct the desired binary solution through the following steps:

\noindent\textbf{Step 1: Analyzing the bilinear constraint.}
The constraint $\bm{W} \diag(\bm{s}) = 0$ implies that for each index $i \in \{1,2,...,n\}$, we have:
\begin{equation}
\hat{\bm{W}}_i \hat{\bm{s}}_i = 0,
\end{equation}
where $\hat{\bm{W}}_i$ denotes the $i$-th column of $\hat{\bm{W}}$. This gives two complementary possibilities for each $i$:
\begin{itemize}
    \item If $\hat{\bm{s}}_i > 0$, then $\hat{\bm{W}}_i$ must be the zero vector.
    \item If $\hat{\bm{W}}_i \neq \bm{0}$, then $\hat{\bm{s}}_i$ must be zero.
\end{itemize}

\noindent\textbf{Step 2: Constructing the binary solution.}
Let $J \subseteq \operatorname{supp}(\hat{\bm{s}})$ be any $\lambda$-element subset of the support of $\hat{\bm{s}}$. We construct a binary vector $\bm{s}'$ as:
\begin{equation}
\bm{s}'_i = \begin{cases} 
1 & \text{if } i \in J, \\
0 & \text{otherwise}.
\end{cases}
\end{equation}
By construction, $\bm{s}'$ satisfies:
\begin{itemize}
    \item $\bm{1}^\top \bm{s}' = \lambda$ (sparsity constraint)
    \item $\bm{s}' \in \{0,1\}^n$ (binary constraint)
    \item $\operatorname{supp}(\bm{s}') = J$
\end{itemize}

\noindent\textbf{Step 3: Verifying feasibility.}
Let $\bm{W}' = \hat{\bm{W}}$. We verify that $(\bm{W}', \bm{s}')$ satisfies all constraints:
\begin{itemize}
    \item For the bilinear constraint (\ref{eq:relaxed-bilinear}):
    \begin{align*}
    \bm{W}' \diag(\bm{s}') &= \hat{\bm{W}} \diag(\bm{s}') \\
    &= \sum_{i=1}^n \hat{\bm{W}}_i \bm{s}'_i \\
    &= \sum_{i \in J} \hat{\bm{W}}_i \cdot 1 + \sum_{i \notin J} \hat{\bm{W}}_i \cdot 0 \\
    &= \sum_{i \in J} \hat{\bm{W}}_i
    \end{align*}
    Since $J \subseteq \operatorname{supp}(\hat{\bm{s}})$, for each $i \in J$ we have $\hat{\bm{s}}_i > 0$, which by the original constraint implies $\hat{\bm{W}}_i = 0$. Thus $\bm{W}' \diag(\bm{s}') = 0$.
    
    \item The sparsity constraint (\ref{eq:relaxed-sparsity}) and binary constraint (\ref{eq:relaxed-binary}) are satisfied by construction of $\bm{s}'$.
\end{itemize}

\noindent\textbf{Step 4: Optimality preservation.}
The objective value remains unchanged since:
\begin{equation}
\frac{1}{2} \|\bm{W}'\bm{X} - \bm{Y}\|_F^2 = \frac{1}{2} \|\hat{\bm{W}}\bm{X} - \bm{Y}\|_F^2,
\end{equation}
and $\hat{\bm{W}}$ was optimal for the relaxed problem. 

Therefore, $(\bm{W}', \bm{s}')$ is indeed an optimal solution to (\ref{eq:structured-relaxed}) with the desired binary property $\bm{s}' \in \{0,1\}^n$ and support $\operatorname{supp}(\bm{s}') = J$.
\end{proof}

%%%%%%%%%%%%%%%%%%%%%%%%%%%%%%%%%%%

\section{Derivations of the W-subproblem in the Penalty Method}\label{W-derivation}
To derive the mentioned closed-form solution to problem~\eqref{equ:W-update}, we express the Frobenius norm and regularization term using trace operations:
$$
\sum_{i=1}^n \bm{s}_i^{(k+1)} \|\bm{W}[:,i]\|_2^2 = \tr(\bm{W}^\top \bm{W} \diag(\bm{s}^{(k+1)})).
$$
Thus, the objective function becomes:
$$
f(\bm{W}) = \frac{1}{2}\|\bm{WX}-\bm{Y}\|_F^2 + \frac{\rho^{(k)}}{2} \tr(\bm{W}^\top \bm{W} \diag(\bm{s}^{(k+1)})).
$$
Taking the gradient yields:
$$
\frac{df}{d\bm{W}} = (\bm{WX} - \bm{Y}) \bm{X}^\top + \rho^{(k)} \bm{W} \diag(\bm{s}^{(k+1)}).
$$
Setting the gradient to zero gives the desired closed-form solution:
$$
\bm{W}^{(k+1)} = \bm{Y} \bm{X}^\top (\bm{X} \bm{X}^\top + \rho^{(k)} \diag(\bm{s}^{(k+1)}))^{-1}.
$$

\end{document}